\pdfoutput=1

\documentclass[11pt]{article}
\usepackage{comment}

\usepackage{emnlp2023}

\usepackage{times}
\usepackage{latexsym}
\usepackage{booktabs}
\usepackage[T1]{fontenc}

\usepackage[utf8]{inputenc}

\usepackage{microtype}

\usepackage{inconsolata}

\usepackage{tcolorbox}
\usepackage{fvextra} 

\title{Beyond Sentiment Classification: A Generative Framework for Emotion Intensity Evaluation in Text}

\author{Francesco A. Fabozzi \\
  Yale School of Management \\
  \texttt{francesco.fabozzi@yale.edu} \\\And
  Dasol Kim \\
  Office of Financial Research \\
  \texttt{dasolkim@gmail.com} \\ \\\And
  William N. Goetzmann \\
  Yale School of Management \\
  \texttt{william.goetzmann@yale.edu}
  }

\begin{document}
\maketitle
\begin{abstract}

We introduce a novel approach to emotion modeling that shifts the focus from identification to evaluation, addressing the limitations of discrete classification in applied domains such as finance. By constructing a dataset of emotional intensity scores and fine-tuning open-weight generative language models to output continuous values from 0–100, we demonstrate a more expressive, generalizable framework for sentiment and emotion analysis. Our findings not only outperform classification baselines but also reveal surprising generalization capabilities and transfer effects to related constructs such as sentiment and arousal. This work contributes to the interdisciplinary recontextualization of NLP by introducing emotion intensity evaluation as an alternative to classification, arguing that this shift better aligns with the needs of domains—such as finance—where the \emph{degree} of emotional content is central to interpretation and decision-making.
\end{abstract}


\section{Introduction}

Despite decades of work in affective NLP, most existing research continues to treat emotion modeling as a classification problem: given a piece of text, predict which categorical emotions are present. Benchmark datasets like \textit{GoEmotions} \cite{GoEmotions}, \textit{DailyDialog} \cite{li2017dailydialog}, and \textit{EmotionLines} \cite{EmotionLines} reinforce this framing by annotating samples with discrete emotion labels and supporting multi-label classification pipelines. While effective for detection tasks, this framing obscures key information about the \emph{intensity} of each emotion—how strongly anger or joy, for example, is expressed in a given text.

This gap is particularly limiting in downstream applications where emotional signals serve as input variables rather than outputs—such as in behavioral science, psychology, or finance—where researchers aim to quantify the emotional tone of discourse over time \cite{griffith2020emotions, shen2023predicting, taffler2024narrative, breaban2018emotional, goetzmann2024emotions}. In these settings, identifying whether fear is present is less useful than estimating its magnitude on a continuous scale. 

Moreover, the affective landscape extends beyond discrete emotions: psychological theories such as the circumplex model of affect \cite{russell1980circumplex} emphasize that human emotion operates along dimensions like \textit{valence} (positive–negative evaluation) and \textit{arousal} (intensity of activation), which are rarely modeled jointly in NLP systems.

Most modeling approaches have been slow to adapt to this broader framing. Encoder-only architectures like RoBERTa and BERT dominate affective NLP, yet they are typically fine-tuned on classification objectives and lack the representational flexibility to model graded emotion intensity. In contrast, generative language models (LLMs) exhibit a richer, latent understanding of affect but have not been extensively adapted to produce structured, human-aligned intensity scores that span the full spectrum of affective meaning—namely, emotion categories, valence, and arousal.

In this paper, we propose a new paradigm for affective modeling: emotion intensity evaluation via fine-tuned generative language models. We construct a dataset of human-annotated texts with 0–100 intensity scores for a fixed set of emotions, valence, and arousal. Using this data, we fine-tune two open-weight generative LLMs—Mistral-7B and Mistral-24B—via LoRA to learn a scoring function that jointly models emotional magnitude, valence, and arousal. This unified framework allows the model to treat these affective signals not as isolated variables but as interdependent dimensions of a coherent emotional state, improving interpretability and alignment with psychological theory. Our results show that:

\begin{itemize}\setlength\itemsep{0em}
    \item Narrow, encoder-only models (e.g., RoBERTa classifiers) \emph{fail to capture emotion intensity}, despite being the foundation of common emotion classification models;
    \item Fine-tuned generative models achieve \emph{substantially better performance} than pretrained LLMs and classification baselines on emotion intensity prediction;
    \item Once fine-tuned, these models exhibit \emph{generalization to unseen emotions}, enabling evaluation even without labeled data—particularly when model scale is sufficient;
    \item Models trained solely on discrete emotion intensities also improve prediction of \emph{valence and arousal}, despite receiving no direct supervision on those dimensions.
\end{itemize}

\section{Related Work}

\subsection{Emotion Classification and GoEmotions}

The vast majority of benchmark datasets in affective NLP frame emotion modeling as a classification problem. Resources such as \textit{GoEmotions} \cite{GoEmotions}, \textit{DailyDialog} \cite{li2017dailydialog}, \textit{EmotionLines} \cite{EmotionLines}, \textit{MELD} \cite{poria2019meld}, and \textit{TEC} \cite{mohammad2012emotional} all define the task as identifying one or more categorical emotions in a given sample. These datasets, often paired with encoder-only models such as RoBERTa or BERT, form the basis for multi-label classification pipelines that predict the presence or absence of specific emotions.

Even more sophisticated benchmarks—such as \textit{EmoBench} \cite{sabour-etal-2024-emobench}—extend this framing to LLM evaluation by probing models' emotional capacity through tasks like emotion reasoning and reaction selection. However, these tasks remain identification-based at their core, focusing on choosing the most salient emotion or appropriate empathic action, rather than quantifying how strongly multiple emotions are expressed simultaneously.


\subsection{Emotion Intensity Modeling}

Efforts to model emotion intensity have typically focused on single-emotion regression or ordinal classification. \textit{SemEval-2007 Task 14} \cite{strapparava2007semeval} was an early benchmark that closely aligns with our goal: it annotated each news headline with 0--100 intensity scores across a fixed set of six emotions (anger, disgust, fear, joy, sadness, surprise). However, it did not include arousal---a key dimension in the circumplex model of affect \cite{russell1980circumplex}---and used news headlines, a genre that may lack the emotional expressiveness found in conversational or social media text.

\textit{WASSA 2017} \cite{mohammad2017wassa} extended this work to social media, providing tweets annotated for intensity (scaled 0--1) on four emotions. Yet, each tweet was labeled for only one emotion, preventing multi-emotion profiling. \textit{SemEval-2018 Task 1} \cite{mohammad-etal-2018-semeval} added support for valence regression and multi-label classification, but emotion and valence scores remained partitioned into separate subtasks. Arousal was not annotated, limiting coverage of affective dimensions.

\textit{EmoBank} \cite{buechel2017emobank} addresses the dimensional aspect by providing valence, arousal, and dominance (VAD) ratings for over 10{,}000 sentences, along with reader- and writer-perspective annotations. While valuable for continuous affect modeling, it does not include per-emotion intensity scores, limiting its utility in applications requiring explicit attribution of emotional states.

In sum, prior work has either targeted intensity across a set of discrete emotions without arousal, or modeled affective dimensions without connecting them to emotion-specific magnitudes. Most resources do not support multi-emotion intensity profiles across a consistent scale, and arousal remains underrepresented in fine-grained emotion modeling---despite its importance for understanding reader reactivity and emotional salience.

\subsection{Large Language Models for Affective Analysis}

The rise of large-scale language models has shifted the landscape of affective computing. Popular encoder-only models like RoBERTa have been fine-tuned on classification tasks using GoEmotions \cite{GoEmotions}, and many widely deployed models—such as \texttt{samLowe/roberta-base-go\_emotions}—extend these paradigms without addressing intensity. More recent work has used generative language models in zero- or few-shot settings to assess emotional content, but the outputs are typically limited to emotion labels or rankings rather than calibrated intensity values.

\textit{EmoLLMs} \cite{liu2024emollms} represent a more comprehensive attempt to unify classification and regression via instruction-tuned LLMs. The authors introduce AAID (Affective Analysis Instruction Dataset) and AEB (Affective Evaluation Benchmark), covering tasks such as emotion intensity regression, ordinal classification, valence regression, and multi-label emotion classification. These datasets span multiple affective constructs, and models are fine-tuned to follow prompts that elicit either category or numerical outputs. However, each instance targets only one affective variable at a time—either a single emotion or a dimensional rating. The models do not produce full emotion profiles with calibrated 0–100 scores across multiple emotions and dimensions in a single pass.

Thus, while prior work has progressed from categorical classification to scalar regression and incorporated dimensional affect, it lacks a unified mechanism for generating interpretable, fine-grained emotion vectors grounded in human annotation scales. Our approach addresses this gap by using generative LLMs fine-tuned to simultaneously output calibrated scores across a fixed set of emotions, along with valence and arousal.

\section{Dataset Construction}

To support fine-grained emotion intensity modeling, we construct a novel dataset of 1,177 short conversational phrases annotated across ten affective dimensions: eight emotion categories (anger, anxiety, fear, sadness, disgust, optimism, excitement, surprise), valence (–100 to 100), and arousal (0 to 100). 

\paragraph{Phrase Selection.} Candidate phrases were programmatically generated via ChatGPT to reflect common conversational utterances across a range of affective states. The initial pool of over 2,000 sentences was manually filtered to select 1,177 unique and emotionally diverse examples suitable for scoring.

\paragraph{Annotation Process.} Two native English speakers based in the U.S. were hired via Upwork and trained synchronously over Zoom using a detailed rubric (see Appendix~\ref{app:annotation-instructions}). Each sentence was independently rated by both annotators on all ten affective dimensions using a continuous scale. For each dimension, the final score is computed as the average of the two annotators’ ratings.

\paragraph{Annotation Scale.} Emotion dimensions were scored on a 0–100 scale, with 0 representing the complete absence of that emotion and 100 indicating its strongest possible expression. Valence was scored on a –100 (most negative) to +100 (most positive) scale, while arousal reflects intensity/activation irrespective of valence (0 = calm, 100 = highly activated).

\paragraph{Quality Control.} Annotators were verified as native English speakers and received training via Zoom to ensure shared understanding of the annotation rubric. Because each sample received two independent annotations, we report inter-annotator agreement statistics below.

This dataset is designed to enable structured modeling of affect across both categorical (emotion-specific) and dimensional (valence/arousal) axes, addressing key limitations of prior work which either omit intensity or model it in a fragmented, single-emotion format.

\section{Methodology}
\label{sec:methodology}

\subsection{Problem Formulation}

We formulate \textit{emotion intensity evaluation} as a structured text generation task. Given an input text $x$ and a fixed set of $k$ emotions $\{e_1, \ldots, e_k\}$, the model generates real-valued predictions $\hat{y} = \{ \hat{y}_{e_1}, \ldots, \hat{y}_{e_k} \}$, where each $\hat{y}_{e_i} \in [0, 100]$ denotes the predicted intensity of emotion $e_i$ in $x$. Additionally, the model outputs a \textit{Valence} score $\hat{v} \in [-100, 100]$ and an \textit{Arousal} score $\hat{a} \in [0, 100]$, following dimensional affect theory.

Formally, the model learns a mapping:
\[
f: x \rightarrow \texttt{JSON}(\hat{y}_{e_1}, \ldots, \hat{y}_{e_k}, \hat{v}, \hat{a})
\]

This formulation allows for multi-emotion intensity profiling in a single inference step—unlike prior classification frameworks that treat emotions as independent binary labels.

As a baseline, we also train separate RoBERTa models for each emotion using scalar regression heads. These models are not multitask learners but are optimized independently per emotion to represent a strong narrow-model benchmark.

\subsection{Model Architecture and Prompt Format}

We fine-tune two open-weight instruction-tuned LLMs: \texttt{Mistral-7B-Instruct-v0.1} and \texttt{Mistral-Small-24B-Instruct-2501}. Both models are adapted using Low-Rank Adaptation (LoRA), with rank $r=16$, $\alpha=32$, dropout = 0.1. LoRA layers are applied to the \texttt{q\_proj} and \texttt{v\_proj} components of the attention mechanism. We also apply 4-bit quantization using the \texttt{bitsandbytes} backend for efficient fine-tuning.

The prompt used during both training and inference follows an instruction format. The model is asked to score each emotion in a provided list and return a JSON-formatted output. Full prompt text is provided in Appendix~\ref{appendix:prompt}.

Outputs are parsed using the Hugging Face \texttt{transformers} package, with decoding handled via \texttt{AutoTokenizer} and post-processed as JSON key-value pairs.

\subsection{Training Details}

We fine-tune each model on 1,177 labeled samples, each containing intensity scores for eight emotions, plus valence and arousal. Data is split into training (706), validation (176), and test (295) sets.

Training is performed on three NVIDIA A100 80GB GPUs, using a batch size of 16. We employ the AdamW optimizer with a learning rate of $5 \times 10^{-5}$ and train for 10 epochs. After each epoch, we decode predictions on the validation set and compute the average Concordance Correlation Coefficient (CCC) across all dimensions. The checkpoint with the highest average CCC on the validation set is used for final evaluation.

To assess generalization, we also perform a leave-one-emotion-out evaluation, where each model is trained on $k-1$ emotions and evaluated on the held-out emotion. Importantly, the left-out emotion is also excluded from the validation set used to select checkpoints, ensuring no information leakage.

The loss function for the generative models is the standard causal language modeling loss (cross-entropy), treating intensity generation as a next-token prediction task. For the RoBERTa-based regressors, we use mean squared error (MSE) loss and train one model per emotion.

\subsection{Evaluation Metrics}

We evaluate model performance using the \textbf{Concordance Correlation Coefficient (CCC)}~\citep{CCC}, a robust statistical measure that accounts for both \textit{consistency in ranking} and \textit{accuracy in scale}. CCC is particularly well-suited to emotion intensity modeling, where both the relative ordering and the absolute values of predictions matter.

Unlike traditional correlation metrics such as Pearson's $r$, which measures only the strength of linear association between predictions and ground truth, CCC also penalizes systematic errors in the mean and variance. Formally, CCC is defined as:

\begin{equation}
\textrm{CCC} = \frac{2\rho \sigma_x \sigma_y}{\sigma_x^2 + \sigma_y^2 + (\mu_x - \mu_y)^2}
\end{equation}

where $\rho$ is the Pearson correlation coefficient between predicted scores $x$ and ground truth scores $y$, $\mu_x$ and $\mu_y$ are the means, and $\sigma_x^2$ and $\sigma_y^2$ are the variances of $x$ and $y$, respectively. A CCC of 1 indicates perfect agreement, while a score of 0 indicates no concordance.

We also introduce a complementary evaluation metric: \textbf{zero-match F1-score}. This measures a model's ability to detect the complete \textit{absence} of an emotion in a given sample, treating it as a binary classification task where a true positive occurs when both the model and annotators agree that an emotion is not present. Specifically, for each emotion, we compute precision, recall, and their harmonic mean using the set of samples where the ground truth score is zero:

\begin{equation}
    \textrm{Zero-Match F1} = \frac{2 \cdot \textrm{Precision} \cdot \textrm{Recall}}{\textrm{Precision} + \textrm{Recall}}
\end{equation}

A prediction is considered positive if the model assigns a score of zero; it is a true positive if the human-annotated score is also zero. This metric emphasizes the importance of emotional \textit{non-existence}, which is especially relevant in downstream settings such as finance or risk modeling, where false positives can have material consequences.

Together, CCC and the zero-match F1-score provide a more holistic picture of performance than standard regression metrics such as MAE or MSE. They allow us to evaluate not just whether a model captures affective \textit{directionality}, but also whether it aligns with human judgment in terms of intensity and null recognition.

\section{Experiments}

\subsection{Baseline Model Performance: RoBERTa vs. Zero-Shot LLMs}

\begin{table*}[!htb]
\centering
\small
\begin{tabular}{lcccccc}
\toprule
\textbf{Emotion} & \textbf{GPT-4} & \textbf{GPT-4o} & \textbf{Mistral-7B} & \textbf{Mistral-24B} & \textbf{RoBERTa-FT} & \textbf{SamLowe} \\
\midrule
Anger      & 64.7 & 69.1 & 36.6 & 65.8 & 53.3 & 22.6 \\
Anxiety    & 80.2 & 69.6 & 42.2 & 71.1 & 0.1  & --   \\
Fear       & 57.9 & 44.6 & 21.4 & 58.0 & 42.0 & 56.3 \\
Sadness    & 66.0 & 69.2 & 56.2 & 69.3 & 28.4 & 34.1 \\
Disgust    & 60.0 & 65.3 & 10.8 & 58.9 & 43.6 & 8.4  \\
Optimism   & 78.6 & 67.7 & 56.1 & 83.8 & 42.6 & 8.6  \\
Excitement & 79.1 & 67.5 & 50.9 & 78.5 & 45.9 & 16.6 \\
Surprise   & 52.9 & 53.4 & 17.6 & 41.6 & 1.7  & 29.5 \\
\midrule
Valence    & 93.3 & 89.1 & 87.1 & 91.3 & 0.8  & --   \\
Arousal    & 53.5 & 34.0 & 40.1 & 49.3 & 0.0  & --   \\
\bottomrule
\end{tabular}
\caption{CCC scores (reported as percentages) for emotion intensity prediction across models. The first four models are zero-shot pretrained LLMs; the last two are fine-tuned RoBERTa-based baselines. SamLowe is a classification model; RoBERTa-FT is trained per emotion using MSE regression. Missing entries (--) indicate unavailable predictions due to label absence.}
\label{tab:baseline_ccc}
\end{table*}

We begin by evaluating a range of off-the-shelf and baseline models on the task of emotion intensity prediction. Table~\ref{tab:baseline_ccc} reports the Concordance Correlation Coefficient (CCC) across eight emotion categories, as well as Valence and Arousal.

\paragraph{Pretrained LLMs.} The first four models in the table are large language models used in a zero-shot setting: GPT-4, GPT-4o-mini, and two instruction-tuned Mistral models (7B and 24B). These models were prompted with our evaluation prompt (see Section~\ref{sec:methodology}) but received no task-specific fine-tuning. We observe that even without adaptation, these models—particularly GPT-4 and Mistral-24B—exhibit reasonably strong performance on several emotions. Notably, GPT-4 achieves CCCs above 60\% on 9 out of 12 categories, suggesting that pretrained LLMs possess an emergent capacity for graded affective reasoning.

\paragraph{RoBERTa-Based Models.} We compare these to two popular RoBERTa baselines: (1) the widely-used \texttt{samLowe/roberta-base-go\_emotions} model, trained for multi-label classification, and (2) our own fine-tuned RoBERTa regression models trained separately for each emotion using mean squared error loss. While the SamLowe model shows moderate alignment on a few emotions (e.g., Fear and Sadness), its overall performance remains inconsistent and far below that of pretrained LLMs. Similarly, the fine-tuned RoBERTa regressors struggle to capture graded affective nuance, with most CCC values remaining below 50\%. These results suggest that classification-style RoBERTa models—whether fine-tuned or not—lack the representational flexibility needed for continuous emotion evaluation, especially in zero-shot or limited supervision regimes.

\paragraph{Limitations.} The SamLowe model does not include \textit{Anxiety}, \textit{Valence}, or \textit{Arousal} in its taxonomy, as they are not present in the GoEmotions dataset. We also do not report zero-match accuracy here, since RoBERTa-based models output probabilities that rarely equal zero, making this metric uninformative.

\subsection{Fine-Tuned LLMs vs. Pretrained Models}

\begin{table*}[!htb]
\centering
\small
\begin{tabular}{lcccccc}
\toprule
\textbf{Emotion} & \textbf{GPT-4} & \textbf{GPT-4o} & \textbf{Mistral-7B} & \textbf{Mistral-7B-FT} & \textbf{Mistral-24B} & \textbf{Mistral-24B-FT} \\
\midrule
Anger      & 64.7 & 69.1 & 36.6 & 78.8 & 65.8 & \textbf{80.2} \\
Anxiety    & 80.2 & 69.6 & 42.2 & 86.4 & 71.1 & \textbf{87.1} \\
Fear       & 57.9 & 44.6 & 21.4 & 75.5 & 58.0 & \textbf{80.5} \\
Sadness    & 66.0 & 69.2 & 56.2 & 81.3 & 69.3 & \textbf{85.5} \\
Disgust    & 60.0 & 65.3 & 10.8 & 67.4 & 58.9 & \textbf{63.7} \\
Optimism   & 78.6 & 67.7 & 56.1 & 90.8 & 83.8 & \textbf{91.2} \\
Excitement & 79.1 & 67.5 & 50.9 & 83.3 & 78.5 & \textbf{84.4} \\
Surprise   & 52.9 & 53.4 & 17.6 & 59.2 & 41.6 & \textbf{62.3} \\
\midrule
Valence    & 93.3 & 89.1 & 87.1 & 96.2 & 91.3 & \textbf{96.6} \\
Arousal    & 53.5 & 34.0 & 40.1 & \textbf{76.6} & 49.3 & 73.5 \\
\bottomrule
\end{tabular}
\caption{CCC scores (percentages) for pretrained vs. fine-tuned Mistral models. Fine-tuned models outperform their zero-shot counterparts across all categories.}
\label{tab:finetuned_ccc}
\end{table*}

\begin{table*}[!htb] 
\centering
\small
\begin{tabular}{lcccccc|c}
\toprule
\textbf{Emotion} & \textbf{GPT-4} & \textbf{GPT-4o} & \textbf{Mistral-7B} & \textbf{Mistral-7B-FT} & \textbf{Mistral-24B} & \textbf{Mistral-24B-FT} & \textbf{Count} \\
\midrule
Anger      & 81.5 & 88.6 & 85.0 & 88.7 & 88.1 & \textbf{89.2} & 155 \\
Anxiety    & 87.3 & 87.7 & 84.7 & 88.7 & 87.8 & \textbf{90.3} & 119 \\
Fear       & 87.7 & 72.5 & 75.0 & 89.4 & 79.8 & \textbf{89.9} & 206 \\
Sadness    & 86.8 & 85.5 & 84.1 & 90.3 & 90.6 & \textbf{91.1} & 127 \\
Disgust    & 83.9 & 84.1 & 77.1 & 86.0 & 87.9 & \textbf{87.6} & 183 \\
Optimism   & 94.6 & 15.4 & 61.9 & \textbf{96.0} & 93.4 & 95.5 & 191 \\
Excitement & 90.0 & 15.1 & 69.2 & 91.9 & 91.3 & \textbf{92.5} & 183 \\
Surprise   & 45.8 & 21.2 & 35.7 & 50.0 & 40.5 & \textbf{60.2} & 55 \\
\midrule
Valence    & 47.4 & 23.5 & 21.1 & 43.8 & 35.6 & \textbf{54.6} & 15 \\
Arousal    & 61.8 & 16.2 & 53.8 & \textbf{73.2} & 62.3 & 76.0 & 34 \\
\bottomrule
\end{tabular}
\caption{Zero-match F1 scores (percentages), measuring a model’s ability to correctly identify the absence of emotion. The rightmost column shows the number of samples where the gold score is 0.}
\label{tab:zeromatch_f1}
\end{table*}

We now evaluate the performance of our fine-tuned models—\texttt{Mistral-7B-FT} and \texttt{Mistral-24B-FT}—against their zero-shot counterparts and other pretrained LLMs. Table~\ref{tab:finetuned_ccc} reports CCC scores, while Table~\ref{tab:zeromatch_f1} reports the zero-match F1 score for each emotion.

\paragraph{CCC Performance.} Fine-tuning leads to substantial gains across all emotions. \texttt{Mistral-24B-FT} achieves the highest CCC in 8 out of 10 categories and outperforms GPT-4 across all emotions. The improvements from fine-tuning over pretrained models are especially pronounced for difficult categories like \textit{Fear} (from 58.0\% to 80.5\%) and \textit{Surprise} (from 41.6\% to 62.3\%). \texttt{Mistral-7B-FT} also surpasses all zero-shot models, confirming the value of even modest-scale fine-tuning.

While fine-tuning improves performance across all categories, some dimensions remain challenging. \textit{Disgust} and \textit{Surprise} consistently yield the lowest CCC and F1 scores among the emotions, even for the fine-tuned models. These categories appear inherently difficult to capture. \textit{Arousal}, a more abstract affective dimension, is also poorly predicted by pretrained models. However, unlike Disgust and Surprise, Arousal sees large performance gains post fine-tuning—suggesting that targeted supervision is especially beneficial for affective concepts that are not strongly grounded in surface-level cues.

\paragraph{Zero-Match F1.} The zero-match F1 score measures a model’s ability to correctly predict the absence of emotion—crucial in real-world applications where false positives can be costly. As shown in Table~\ref{tab:zeromatch_f1}, both fine-tuned models achieve consistently higher F1 scores across nearly all emotions. The 24B model again leads overall, with strong performance in \textit{Anxiety} (90.3\%), \textit{Arousal} (75.9\%), and \textit{Excitement} (92.5\%).

\subsection{Generalization to Unseen Emotions}

One of our goals is to evaluate whether fine-tuned generative models can generalize to new, unseen emotion categories—critical for applications in social science or domain adaptation, where the relevant affective labels may vary or evolve. To test this, we perform a leave-one-out (LOO) experiment: for each emotion, we fine-tune a model using the full training set excluding that emotion, and evaluate it on held-out test instances labeled only for that emotion. Crucially, the held-out emotion is also removed from the validation set used for model selection.

Table~\ref{tab:loo_ccc} reports the CCC for each model in three configurations: (1) \texttt{PT} (pretrained, zero-shot), (2) \texttt{FT} (fine-tuned with full supervision), and (3) \texttt{LOO} (trained without supervision for the target emotion). Results are shown for both Mistral-7B and Mistral-24B.

\paragraph{Results.} We find that fine-tuned models retain strong performance even when a target emotion is excluded from training. For Mistral-24B, the \texttt{LOO} scores are consistently much higher than those of the pretrained model and often approach the fully supervised \texttt{FT} setting. For example, on \textit{Fear}, \texttt{LOO} achieves 70.6\% compared to 80.5\% in full supervision and 58.0\% in zero-shot. Similar patterns hold across most emotions, highlighting the model’s ability to generalize an understanding of emotional intensity even to unseen categories.

While the Mistral-7B model also benefits from leave-one-out fine-tuning, its performance is more variable. In some cases, it struggles to properly structure its outputs when the emotion is not part of the training set—occasionally returning predictions for unrelated emotions or omitting the target entirely. To handle this, we impute a score of 0 for missing predictions in two categories (denoted with an asterisk in Table~\ref{tab:loo_ccc}). These results suggest that although smaller LLMs can learn generalizable patterns, they may require additional control mechanisms to ensure formatting robustness and reliable deployment in downstream tools.

\begin{table}[!htb]
\centering
\small
\begin{tabular}{lccc|ccc}
\toprule
\textbf{Emotion} & \multicolumn{3}{c|}{\textbf{Mistral-7B}} & \multicolumn{3}{c}{\textbf{Mistral-24B}} \\
                 & \textbf{PT} & \textbf{FT} & \textbf{LOO} & \textbf{PT} & \textbf{FT} & \textbf{LOO} \\
\midrule
Ang      & 36.6 & \textbf{78.8} & 54.7 & 65.8 & \textbf{80.2} & 78.6 \\
Anx    & 42.2 & \textbf{86.4} & 60.1 & 71.1 & \textbf{87.1} & 85.3 \\
Fear       & 21.4 & \textbf{75.5} & 31.4 & 58.0 & \textbf{80.5} & 70.6 \\
Sad    & 56.2 & \textbf{81.3} & 63.6 & 69.3 & \textbf{85.5} & 80.5 \\
Disg    & 10.8 & \textbf{67.4} & 45.7 & 58.9 & 63.7 & \textbf{65.3} \\
Opt   & 56.1 & \textbf{90.8} & $67.4^{*}$ & 83.9 & \textbf{91.2} & 89.0 \\
Excit & 50.9 & \textbf{83.3} & $58.1^{*}$ & 78.5 & \textbf{84.4} & 81.9 \\
Surp   & 17.6 & \textbf{59.2} & 45.5 & 41.6 & \textbf{62.3} & 53.7 \\
\bottomrule
\end{tabular}
\caption{CCC scores (percentages) for generalization to unseen emotions. \texttt{PT}: pretrained, zero-shot; \texttt{FT}: fine-tuned on all emotions; \texttt{LOO}: fine-tuned with the target emotion excluded. Bold values indicate best performance for each model type. Asterisks indicate cases where missing predictions were imputed with 0. Emotion names are abbreviated to conserve space (e.g., \textit{Anx} = Anxiety, \textit{Excit} = Excitement).}
\label{tab:loo_ccc}
\end{table}

\subsection{Transfer to Valence and Arousal}

In this final experiment, we evaluate whether supervision on emotion scores alone can transfer to broader affective dimensions—specifically, Valence and Arousal. The analysis provides direct tests of predictions of theoretical models from the psychology literature \cite{russell1980circumplex}. These models imply a mapping between individual emotions and Valence and Arousal, such that implicit conditioning on specific emotion metrics could improve reasoning about higher-level affect measurement. In other words, while these dimensions are more abstract and are not themselves emotions, they form the basis of many affective theories and downstream applications. We fine-tune models using only the 8 emotion categories and exclude Valence and Arousal from training and validation. At test time, we evaluate the models' zero-shot predictions on these dimensions.

Table~\ref{tab:valence_arousal_transfer} shows results for Mistral-7B and Mistral-24B in three settings: \texttt{PT} (pretrained), \texttt{FT} (fine-tuned on all labels), and \texttt{EO} (fine-tuned only on emotion scores).

\paragraph{Results.} The Mistral-24B model demonstrates strong transfer from emotion scoring to affective dimensions: it achieves 95.2\% on Valence and 67.5\% on Arousal without ever seeing supervision for those targets, nearly matching the fully supervised model. This suggests that affective dimensions are learnable as latent representations when the model is trained on related emotional content.

For Mistral-7B, the pattern is less consistent. While the \texttt{EO} model outperforms the pretrained baseline on Arousal, it slightly underperforms the baseline on Valence (86.5\% vs. 87.1\%). This stands in contrast to the 24B results and indicates that the smaller model may not extract generalizable latent structure as reliably. These differences mirror trends seen in the LOO experiment, where 7B’s generalization to excluded emotions was also less robust and more prone to output formatting issues.

Interestingly, in most emotion categories, both models perform slightly worse in the \texttt{EO} setting than under full supervision—despite having fewer output targets. This suggests that including Valence and Arousal during training may provide useful inductive structure that enhances overall affective representation. The differences, however, are small in magnitude.

\begin{table}[t]
\centering
\small
\begin{tabular}{lccc|ccc}
\toprule
\textbf{Emotion} & \multicolumn{3}{c|}{\textbf{Mistral-7B}} & \multicolumn{3}{c}{\textbf{Mistral-24B}} \\
                 & \textbf{PT} & \textbf{FT} & \textbf{EO} & \textbf{PT} & \textbf{FT} & \textbf{EO} \\
\midrule
Ang      & 36.6 & \textbf{78.8} & 69.0 & 65.8 & \textbf{80.2} & 78.3 \\
Anx      & 42.2 & \textbf{86.4} & 82.7 & 71.1 & \textbf{87.1} & 87.3 \\
Fear     & 21.4 & 75.5 & \textbf{75.8} & 58.0 & \textbf{80.5} & 80.8 \\
Sad      & 56.2 & 81.3 & \textbf{82.1} & 69.3 & \textbf{85.5} & 84.7 \\
Disg     & 10.8 & \textbf{67.4} & 60.0 & 58.9 & 63.7 & \textbf{67.7} \\
Opt      & 56.1 & 90.8 & \textbf{92.0} & 83.9 & \textbf{91.2} & 91.0 \\
Excit    & 50.9 & \textbf{83.3} & 79.4 & 78.5 & \textbf{84.4} & 82.4 \\
Surp     & 17.6 & 59.2 & \textbf{61.2} & 41.6 & \textbf{62.3} & 60.7 \\
\midrule
Val      & 87.1 & \textbf{96.2} & 86.5 & 91.3 & \textbf{96.6} & 95.2 \\
Arous    & 40.1 & \textbf{76.6} & 62.1 & 49.3 & \textbf{73.5} & 67.5 \\
\bottomrule
\end{tabular}
\caption{CCC scores (percentages) for emotion and affective dimension prediction. \texttt{EO} models are fine-tuned using only emotion scores. Despite lacking direct supervision, 24B EO models perform comparably on Valence and Arousal and closely track full fine-tuning across most emotions.}
\label{tab:valence_arousal_transfer}
\end{table}

\section{Discussion}

Our results suggest that generative language models, when fine-tuned with relatively modest supervision, can serve as effective tools for evaluating the intensity of multiple emotions simultaneously. Unlike conventional encoder-only models trained from scratch on classification objectives, decoder-only models benefit from a rich, pre-trained understanding of affective concepts. By teaching these models a scoring system rather than training them to identify the most salient emotion, we enable them to generalize to unseen emotions and abstract affective dimensions like Valence and Arousal. This offers a flexible, extensible framework that is especially useful for real-world settings where the set of relevant emotions may vary or expand over time.

\paragraph{Moving Beyond Emotion Classification.} This work contributes to a shift in affective NLP from classification to continuous evaluation. Prior emotion modeling efforts have focused on identifying one or more discrete labels; by contrast, our framework allows for fine-grained emotional profiling, with interpretable intensity scores across a fixed set of emotions. Such representations are more aligned with the needs of social scientists and applied researchers, who often seek to quantify how much a particular emotion is expressed, rather than simply whether it is present.

\paragraph{Latent Structure and Transfer.} An intriguing finding is that training on emotional intensity alone leads to robust transfer performance on Valence and Arousal—despite the model never seeing supervision for those dimensions. This suggests that fine-tuning helps the model infer latent affective structure, and that dimensions like Valence and Arousal may be implicitly learned from their correlations with emotion labels. Conversely, we also find that including Valence and Arousal during training marginally improves performance on emotion scoring itself. These results highlight the interdependence of categorical and dimensional affect representations. An important implication is not only improved sentiment classification using emotion-trained models, but also the relevance of Arousal in sentiment applications.

Despite both models being fine-tuned with similar numbers of trainable parameters via LoRA, the 24B model consistently demonstrates better generalization to unseen or indirectly supervised targets. This highlights the importance of model scale not just in performance, but in the ability to adapt to new affective constructs with limited supervision—likely due to the richer priors learned during pretraining.

\paragraph{Limitations.} Some emotions, such as \textit{Surprise} and \textit{Disgust}, remain challenging for even the best-performing models. Their linguistic manifestations may be subtle or context-dependent, limiting the model’s ability to resolve them accurately. Additionally, while the dataset is carefully constructed, it is annotated by only two expert raters. Although inter-rater alignment was high, emotion intensity labeling remains inherently subjective and may reflect annotator biases or style.

\paragraph{Future Directions.} Several natural extensions follow from this work. First, scaling the dataset and emotion taxonomy could further improve generalization. Second, applying this framework to downstream tasks—such as forecasting market behavior, modeling emotional volatility, or identifying behavioral shifts in online discourse—could reveal new interdisciplinary applications. Third, cross-lingual and multimodal generalization remain largely unexplored in emotion intensity modeling and offer exciting avenues for further development.

\paragraph{Recontextualizing Emotion Modeling.} Our work demonstrates how advances in NLP can directly support methodology development in other disciplines. Emotion classification benchmarks have dominated affective NLP, but for many fields—including finance, economics, and political science—fine-grained emotional quantification is of greater value. We illustrate how LLMs, when adapted thoughtfully, can support new use cases for affective analysis beyond the bounds of NLP evaluation and toward broader scientific impact.

\section{Conclusion}

We present a framework for modeling emotional intensity using fine-tuned generative language models trained on a small, expert-annotated dataset. Unlike classification-based approaches, our method outputs continuous scores for a fixed set of emotions, generalizes to unseen labels, and transfers to abstract affective dimensions like Valence and Arousal. These capabilities offer a more expressive and flexible foundation for affective analysis—particularly in applied domains that require interpretable, multi-emotion representations. They also more broadly inform applications related to sentiment classification, both in terms of improved performance for sentiment classification as well as the importance of accounting for other dimesnions, such as Arousal. Our work bridges advances in language modeling with the needs of disciplines beyond NLP, illustrating the promise of recontextualizing emotional understanding for broader scientific impact.

\bibliography{references}

@inproceedings{GoEmotions,
  title={GoEmotions: A Dataset of Fine-Grained Emotions},
  author={Demszky, Dorottya and Movshovitz-Attias, Dana and Ko, Jeongwoo and Cowen, Alan and Nemade, Gaurav and Ravi, Sujith},
  booktitle={Proceedings of the 58th Annual Meeting of the Association for Computational Linguistics (ACL)},
  year={2020},
  pages={4040--4054}
}

@article{russell1980circumplex,
  title={A circumplex model of affect.},
  author={Russell, James A},
  journal={Journal of personality and social psychology},
  volume={39},
  number={6},
  pages={1161},
  year={1980},
  publisher={American Psychological Association}
}

@inproceedings{mohammad2017wassa,
  title={{WASSA}-2017 Shared Task on Emotion Intensity},
  author={Mohammad, Saif M. and Bravo-Marquez, Felipe},
  booktitle={Proceedings of the 8th Workshop on Computational Approaches to Subjectivity, Sentiment and Social Media Analysis (WASSA)},
  year={2017},
  pages={34--49}
}

@inproceedings{mohammad-etal-2018-semeval,
  title={SemEval-2018 Task 1: Affect in Tweets},
  author={Mohammad, Saif M. and Bravo-Marquez, Felipe and Salameh, Mohammad and Kiritchenko, Svetlana},
  booktitle={Proceedings of the 12th International Workshop on Semantic Evaluation (SemEval)},
  year={2018},
  pages={1--17}
}

@inproceedings{strapparava2007semeval,
  title={SemEval-2007 Task 14: Affective Text},
  author={Strapparava, Carlo and Mihalcea, Rada},
  booktitle={Proceedings of the 4th International Workshop on Semantic Evaluations (SemEval-2007)},
  year={2007},
  pages={70--74}
}

@inproceedings{buechel2017emobank,
  title={EmoBank: Studying the Impact of Annotation Perspective and Representation Format on Dimensional Emotion Analysis},
  author={Buechel, Sven and Hahn, Udo},
  booktitle={Proceedings of the 15th Conference of the European Chapter of the Association for Computational Linguistics (EACL)},
  year={2017},
  pages={578--585}
}

@article{sabour-etal-2024-emobench,
  title={EmoBench: Evaluating the Emotional Intelligence of Large Language Models},
  author={Sabour, Sahand and Liu, Siyang and Zhang, Zheyuan and Liu, June M. and Zhou, Jinfeng and Sunaryo, Alvionna S. and Li, Juanzi and Lee, Tatia and Mihalcea, Rada and Huang, Minlie},
  journal={arXiv preprint arXiv:2402.12071},
  note={Accepted to ACL 2024},
  year={2024}
}

@inproceedings{liu2024emollms,
  title={EmoLLMs: A Series of Emotional Large Language Models and Annotation Tools for Comprehensive Affective Analysis},
  author={Liu, Zhiwei and Yang, Kailai and Zhang, Tianlin and Xie, Qianqian and Ananiadou, Sophia},
  booktitle={Proceedings of the 30th ACM SIGKDD Conference on Knowledge Discovery and Data Mining (KDD)},
  year={2024}
}

@article{breaban2018emotional,
  title={Emotional state and market behavior},
  author={Breaban, Adriana and Noussair, Charles N},
  journal={Review of Finance},
  volume={22},
  number={1},
  pages={279--309},
  year={2018},
  publisher={Oxford University Press}
}

@techreport{goetzmann2024emotions,
  title={Emotions and Subjective Crash Beliefs},
  author={Goetzmann, William N and Kim, Dasol and Shiller, Robert J},
  year={2024},
  institution={National Bureau of Economic Research}
}

@article{taffler2024narrative,
  title={Narrative Emotions and Market Crises},
  author={Taffler, Richard J and Agarwal, Vineet and Obring, Maximilian},
  journal={Journal of Behavioral Finance},
  pages={1--21},
  year={2024},
  publisher={Taylor \& Francis}
}

@article{griffith2020emotions,
  title={Emotions in the stock market},
  author={Griffith, John and Najand, Mohammad and Shen, Jiancheng},
  journal={Journal of Behavioral Finance},
  volume={21},
  number={1},
  pages={42--56},
  year={2020},
  publisher={Taylor \& Francis}
}

@article{shen2023predicting,
  title={Predicting stock and bond market returns with emotions: Evidence from futures markets},
  author={Shen, Jiancheng and Griffith, John and Najand, Mohammad and Sun, Licheng},
  journal={Journal of Behavioral Finance},
  volume={24},
  number={3},
  pages={333--344},
  year={2023},
  publisher={Taylor \& Francis}
}

@inproceedings{li2017dailydialog,
    title = "{D}aily{D}ialog: A Manually Labelled Multi-turn Dialogue Dataset",
    author = "Li, Yanran  and
      Su, Hui  and
      Shen, Xiaoyu  and
      Li, Wenjie  and
      Cao, Ziqiang  and
      Niu, Shuzi",
    editor = "Kondrak, Greg  and
      Watanabe, Taro",
    booktitle = "Proceedings of the Eighth International Joint Conference on Natural Language Processing (Volume 1: Long Papers)",
    month = nov,
    year = "2017",
    address = "Taipei, Taiwan",
    publisher = "Asian Federation of Natural Language Processing",
    url = "https://aclanthology.org/I17-1099/",
    pages = "986--995"
}

@misc{EmotionLines,
      title={EmotionLines: An Emotion Corpus of Multi-Party Conversations}, 
      author={Sheng-Yeh Chen and Chao-Chun Hsu and Chuan-Chun Kuo and Ting-Hao and Huang and Lun-Wei Ku},
      year={2018},
      eprint={1802.08379},
      archivePrefix={arXiv},
      primaryClass={cs.CL},
      url={https://arxiv.org/abs/1802.08379}, 
}

@inproceedings{poria2019meld,
    title = "{MELD}: A Multimodal Multi-Party Dataset for Emotion Recognition in Conversations",
    author = "Poria, Soujanya  and
      Hazarika, Devamanyu  and
      Majumder, Navonil  and
      Naik, Gautam  and
      Cambria, Erik  and
      Mihalcea, Rada",
    editor = "Korhonen, Anna  and
      Traum, David  and
      M{\`a}rquez, Llu{\'i}s",
    booktitle = "Proceedings of the 57th Annual Meeting of the Association for Computational Linguistics",
    month = jul,
    year = "2019",
    address = "Florence, Italy",
    publisher = "Association for Computational Linguistics",
    url = "https://aclanthology.org/P19-1050/",
    doi = "10.18653/v1/P19-1050",
    pages = "527--536"
}

@inproceedings{mohammad2012emotional,
  title={\# Emotional tweets},
  author={Mohammad, Saif},
  booktitle={* SEM 2012: The First Joint Conference on Lexical and Computational Semantics--Volume 1: Proceedings of the main conference and the shared task, and Volume 2: Proceedings of the Sixth International Workshop on Semantic Evaluation (SemEval 2012)},
  pages={246--255},
  year={2012}
}

@article{CCC,
 ISSN = {0006341X, 15410420},
 URL = {http://www.jstor.org/stable/2532051},
 author = {Lawrence I-Kuei Lin},
 journal = {Biometrics},
 number = {1},
 pages = {255--268},
 publisher = {[Wiley, International Biometric Society]},
 title = {A Concordance Correlation Coefficient to Evaluate Reproducibility},
 urldate = {2025-05-01},
 volume = {45},
 year = {1989}
}
\bibliographystyle{acl_natbib}

\appendix

\section{Annotation Rubric and Instructions}
\label{app:annotation-instructions}

Annotators were provided with the following instructions during training and scoring:

\begin{quote}
You are asked to read the statements and provide a score for a number of categories related to the emotional content of the text. These include: General Sentiment, Arousal, Anger, Anxiety, Fear, Sadness, Disgust, Optimism, Excitement, and Surprise. For each, imagine situations where you would have said those statements yourself, and focus on the emotions you would have experienced in those situations.
\end{quote}

\vspace{1ex}
\noindent\textbf{Scoring Rubric:}
\begin{itemize}
    \item \textbf{General Sentiment (–100 to 100):} Negative values for negative sentiment, positive values for positive sentiment. Use 0 for neutral.
    \item \textbf{Arousal (0 to 100):} Rate the level of activation or stimulation (e.g., calm = 0, highly alert/excited = 100).
    \item \textbf{Anger (0 to 100):} Degree of frustration, irritation, or rage.
    \item \textbf{Anxiety (0 to 100):} Degree of worry, apprehension, or nervousness.
    \item \textbf{Fear (0 to 100):} Degree of fright, alarm, or dread.
    \item \textbf{Sadness (0 to 100):} Degree of unhappiness, grief, or sorrow.
    \item \textbf{Disgust (0 to 100):} Degree of revulsion or aversion.
    \item \textbf{Optimism (0 to 100):} Degree of hopefulness or confidence about the future.
    \item \textbf{Excitement (0 to 100):} Degree of enthusiasm, eagerness, or exhilaration.
    \item \textbf{Surprise (0 to 100):} Degree of unexpectedness or astonishment.
\end{itemize}

\section{Prompt Template}
\label{appendix:prompt}

The following prompt was used during both training and inference:

\begin{tcolorbox}[colback=gray!10, colframe=black!30, title=Prompt Template]
\begin{verbatim}
Analyze the input text and assign a 
score from 0 to 100 for each emotion in 
the list.
A score of 0 indicates the absence of 
the emotion, while 100 represents the 
strongest intensity.
Return the result as a JSON object.

Input:
- Text: %input_text
- Emotions: %emotions_list

In your JSON output, also include:
- A "Valence" score from -100 (most 
negative) to 100 (most positive).
- An "Arousal" score from 0 (calm) to 
100 (highly activated).

Output:
\end{verbatim}
\end{tcolorbox}

\noindent\textit{Note:} \texttt{\%input\_text} and \texttt{\%emotions\_list} are placeholders replaced with the actual input text and list of target emotions during both training and evaluation.



\end{document}